\documentclass[conference]{IEEEtran}

\usepackage[utf8]{inputenc}
\usepackage[T1]{fontenc}
\usepackage{microtype}
\usepackage[ngerman,english]{babel}
\usepackage{listings} 
\usepackage{amsmath}
\usepackage[hidelinks]{hyperref}
\usepackage{cleveref}
\usepackage{multicol}
\usepackage{glossaries}
\usepackage[commandnameprefix=always,todonotes={textsize=tiny,obeyFinal}]{changes}
\usepackage{tikz}
\usepackage[binary-units=true]{siunitx}
\usepackage{dblfloatfix}
\usepackage{bm} 
\usepackage{mathtools}
\usepackage{physics}
\usepackage{graphicx}
\graphicspath{{./fig/}}
\usepackage{booktabs}
\usepackage{dcolumn} 
\newcolumntype{C}{D{,}{,\,}{2.2}}

\usepackage[backend=biber,maxbibnames=1,minbibnames=1,mincitenames=1,firstinits=true,sorting=none,style=ieee,doi=true,eprint=false,isbn=false,url=false]{biblatex}
\newacronym{adc}{ADC}{analog-to-digital converter}
\newacronym{adex}{AdEx}{adaptive exponential integrate-and-fire}
\newacronym{bss2}{\mbox{BSS-2}}{Brain\mbox{ScaleS-2}}
\newacronym{cmos}{CMOS}{complementary metal-oxide semiconductor}
\newacronym{maf}{MAF}{masked autoregressive flow}
\newacronym{nde}{NDE}{neural density estimator}
\newacronym{sbi}{SBI}{simulation-based inference}
\newacronym{snpe}{SNPE}{sequential neural posterior estimation}
\newacronym{vae}{VAE}{variational autoencoder}

\newcommand{\myvec}[1]{\bm{#1}}
\newcommand{\captiontitle}[1]{\textbf{#1 -- }}

\makeatletter
\newcommand{\@giventhatstar}[2]{\left(#1\;\middle|\;#2\right)}
\newcommand{\@giventhatnostar}[3][]{#1(#2\;#1|\;#3#1)}
\newcommand{\giventhat}{\@ifstar\@giventhatstar\@giventhatnostar}
\makeatother

\begin{document}

\title{Reproduction of AdEx dynamics on neuromorphic hardware through data embedding and simulation-based inference}

\author{%
	\IEEEauthorblockN{%
		Jakob Huhle\IEEEauthorrefmark{1}, Jakob Kaiser\IEEEauthorrefmark{1}\IEEEauthorrefmark{2}, Eric Müller\IEEEauthorrefmark{1} and Johannes Schemmel\IEEEauthorrefmark{1}}%
	\IEEEauthorblockA{\IEEEauthorrefmark{1}\textit{Kirchhoff Institute for Physics}, Heidelberg University, Germany}%
	\IEEEauthorblockA{\IEEEauthorrefmark{2}\href{mailto:Jakob.Kaiser@kip.uni-heidelberg.de}{Jakob.Kaiser@kip.uni-heidelberg.de}}%
}

\maketitle

\begin{abstract}
The development of mechanistic models of physical systems is essential for understanding their behavior and formulating predictions that can be validated experimentally.
Calibration of these models, especially for complex systems, requires automated optimization methods due to the impracticality of manual parameter tuning.
In this study, we use an autoencoder to automatically extract relevant features from the membrane trace of a complex neuron model emulated on the \acrlong{bss2} neuromorphic system, and subsequently leverage \gls{snpe}, a \acrlong{sbi} algorithm,
to approximate the posterior distribution of neuron parameters.

Our results demonstrate that the autoencoder is able to extract essential features from the observed membrane traces, with which the \gls{snpe} algorithm is able to find an approximation of the posterior distribution.
This suggests that the combination of an autoencoder with the \gls{snpe} algorithm is a promising optimization method for complex systems.

\end{abstract}


\section{Introduction}\label{sec:introduction}
\glsresetall

In science, researchers endeavor to develop mechanistic models for physical systems.
Subsequently, these models can be utilized to comprehend the behavior of the system and to formulate predictions that can be subjected to experimental verification.
Once a model candidate has been identified, the model parameters must be calibrated to ensure that the model can reproduce the observed behavior of the system.

In the case of complex models, manual tuning of the parameters is not a viable option; instead, automated optimization methods are required \cite{vanier1999comparative}.
Genetic algorithms have been demonstrated to be an effective approach for identifying parameters in complex models within the field of neuroscience \cite{gouwens2018systematic, druckmann07_nourl, vanier1999comparative}.
Gradient-based optimization methods are capable of directed optimization, which makes them potentially more efficient for finding suitable model parameters \cite{deistler2024differentiable}.
\Gls{sbi} methods can be employed to approximate the posterior distribution of model parameters, thereby providing additional insight into the sensitivity and correlation of the parameters \cite{cranmer2020frontier, lueckmann2017flexible, greenberg2019automatic, goncalves2020training, deistler2022truncated}.

In previous applications of the \gls{snpe} algorithm, an \gls{sbi} algorithm, to neuroscientific problems, experiments relied on handcrafted features which were extracted from the recordings of neural traces \cite{goncalves2020training, kaiser2023simulation}.
In the present study, we utilize an autoencoder to extract relevant features from the membrane trace of a complex neuron model emulated on the \gls{bss2} neuromorphic system \cite{pehle2022brainscales2, kaiser2023simulation} and subsequently employ the \gls{snpe} algorithm to approximate the parameters of the neuron model.
It is our hope that the automatic feature extraction facilitated by autoencoders will prove instrumental in enabling the application of \gls{sbi} methods to more complex problems, thereby alleviating the need for handcrafted features.

\subsection{The \acrlong{bss2} System}\label{sec:intro:bss2}
\Gls{bss2} is a mixed-signal neuromorphic system;
while synapse and neuron dynamics are emulated in analog circuits, spike communication and configuration is handled digitally \cite{pehle2022brainscales2,billaudelle2022accurate}.
The system provides \num{512} analog neuron circuits which emulate the dynamics of the \gls{adex} \cite{brette2005adaptive} neuron model in continuous time \cite{billaudelle2022accurate}:
\begin{equation}
	\label{eq:adex}
	\begin{split}
		C_\text{m} \dv{V_\text{m}(t)}{t} & = g_\text{L} \cdot \left( V_\text{L} - V_\text{m}(t) \right) \\
		                                 & + g_\text{L} \Delta_\text{T} \exp\left( \frac{V_\text{m}(t) - V_\text{T}}{\Delta_\text{T}} \right) \\
										 & + I_\text{syn}(t) + I(t) - w(t),
	\end{split}
\end{equation}
where $V_\text{m}$ is the membrane potential, $C_\text{m}$ the membrane capacitance, $V_\text{L}$ the leak potential, $g_\text{L}$ the leak conductance, $\Delta_\text{T}$ the threshold slope factor and $V_\text{T}$ the effective threshold potential.

$I_\text{syn}(t)$ describes the synaptic current and will not be used in this study; $I(t)$ is an arbitrary current injected on the membrane.
$w(t)$ represents an adaptation current with the following dynamics
\begin{equation}
	\label{eq:adaptation}
		\tau_\text{w} \dv{w(t)}{t} = a \left( V_\text{m}(t) - V_\text{L} \right) - w(t),
\end{equation}
here $a$ is the subthreshold adaptation and $\tau_\text{w} = \frac{C_\text{w}}{g_{\tau_\text{w}}}$ the adaptation time constant which is determined by the ratio of the capacitance $C_\text{w}$ and the conductance $g_{\tau_\text{w}}$.

As soon as the membrane potential $V_\text{m}$ reaches the threshold potential $V_\text{th}$, the membrane potential is set to the reset potential $V_\text{r}$ and the adaptation current is increased by the spike-triggered adaptation $b$: $w \rightarrow w + b$.
The membrane potential $V_\text{m}$ is kept at the reset potential $V_\text{r}$ during the refractory period $\tau_\text{ref}$; afterwards it evolves according to \cref{eq:adex} again.

The circuits are implemented in \SI{65}{\nm} \gls{cmos} technology.
Compared to the biological time domain, the neuron dynamics evolve in accelerated time with a tunable speed; in this study, we choose a speed-up factor of \num{1000}.

The parameters of each neuron can be configured individually using a capacitor-based memory array \cite{hock13analogmemory};
the array is configured using digital \SI{10}{\bit} values and provides analog currents and voltages which control the behavior of the neuron\footnote{The last value is reserved such that the parameters are adjustable in a range from \numrange{0}{1022}.}.

\subsection{Simulation-based Inference}
\Gls{sbi} offers the possibility to approximate the posterior distribution of model parameters even if the likelihood is not tractable, i.e.\ it can not be calculated or is too expensive to calculate.
More precisely, given a model $\mathcal{M}: \Theta \rightarrow X$ which is configured by the parameters $\myvec{\theta} \in \Theta$ and creates observations $\myvec{x} \in X$, \gls{sbi} can be used to approximate the posterior distribution $p\giventhat{\myvec{\theta}}{\myvec{x}^*}$ for a given target observation $\myvec{x}^* \in X$.
In this manuscript we will use the \gls{snpe} algorithm \cite{papamakarios2016fast,lueckmann2017flexible,greenberg2019automatic} to approximate the posterior distribution of \gls{adex} parameters.
The algorithm takes a mechanistic model $\mathcal{M}$, a prior distribution of the model parameters $p(\myvec{\theta})$ and a target observation $\myvec{x}^*$ as an input.

As a first step, random parameters $\myvec{\theta}' \sim p(\myvec{\theta})$ are drawn from the prior.
These parameters are injected in the model $\mathcal{M}$ to produce observations $\{\myvec{x}'\}_i$.
Therefore, we implicitly sample from the likelihood $\myvec{x}' \sim p\giventhat{\myvec{x}}{\myvec{\theta}'}$.

The random parameters and observations $\{\myvec{\theta}', \myvec{x}'\}_i$ are then used to train a \gls{nde}\footnote{\Glspl{nde} are flexible sets of probability distributions which are configured by neural networks; we will use a \gls{maf} \cite{papamakarios2017masked, papamakarios2019sequential, goncalves2020training} as an \gls{nde}.} which approximates the posterior distribution $p\giventhat{\myvec{\theta}}{\myvec{x}}$.
During training the negative log-likelihood of the posterior density estimate of the drawn samples $\{\myvec{\theta}', \myvec{x}'\}_i$ is minimized and the \gls{nde} learns to approximate the posterior distribution $p\giventhat{\myvec{\theta}}{\myvec{x}}$ for any observation $\myvec{x}$.

This posterior can then be used to draw additional samples for a given target observation $\myvec{x}^*$ and to train the \gls{nde} again to improve the approximation of the posterior for the given observation\footnote{After this step the posterior approximation is no longer amortized, i.e.\ it cannot be used to infer parameters for any observation $\myvec{x}$ but only for the target observation $\myvec{x}^*$.}.
This step can be repeated several times to further improve the approximation.
For further details on the \gls{snpe} algorithm and its applications see \cite{lueckmann2017flexible,greenberg2019automatic,goncalves2020training,kaiser2023simulation}.

\subsection{Autoencoder}
High-dimensional data often contains redundant or irrelevant information for the task at hand.
In deep learning, this can necessitate higher model complexity, prolong training times, and increase the risk of overfitting, which leads to worse performance in generalization on new, unseen data.
Therefore, dimensionality reduction techniques can be employed to learn a lower-dimensional representation of the original data.

An autoencoder is an unsupervised deep learning model that can be leveraged for such a dimensionality reduction task.
It consists of two neural networks: an encoder and a decoder.
The encoder transforms the data into a lower-dimensional feature space, referred to as the latent space.
Conversely, the decoder then maps the features from the latent space back to the original input space, aiming to recover the original input.
The autoencoder is thus trained by minimizing a distance metric, quantifying the difference between the original data and its reconstruction.
Once the model is successfully trained, the encoder can be used independently to obtain a lower-dimensional representation for each data sample.

For more information on autoencoders, see \cite{li2023comprehensive, yildirim2018efficient}.

\section{Methods}\label{sec:methods}
We first introduce the experimental routine with which we will record the behavior of an \gls{adex} neuron on \gls{bss2}.
Next, we explain how we create a dataset which will then be used to train an autoencoder.
Finally, we employ that trained autoencoder to extract relevant features from the experiment observation and utilize the \gls{snpe} algorithm to approximate the posterior distribution of model parameters.

\subsection{Experimental Setup}\label{sec:meth:setup}
Similar to \cite{naud2008firing}, we investigate the behavior of a single \gls{adex} neuron when stimulated with a step current $I(t)$, compare \cref{eq:adex}.
We emulate the dynamics on the \gls{bss2} system and record the membrane potential $V_\text{m}$ for \SI{1}{\ms}\footnote{Due to the \num{1000}-fold speed up of the \gls{bss2} system, this corresponds to \SI{1}{\s} in biological time.} after the step current was enabled.
The recorded membrane trace is then down sampled to \num{10000} data points to reduce memory consumption and computational cost in subsequent steps.
We manually chose parameters which will be used to create a target observation $\myvec{x}^*$.
Four of these parameters will later be altered to create a dataset and to see if the \gls{snpe} algorithm is able to find an approximation of their posterior distribution: the adaptation parameters $a$, $b$ and $g_{\tau_\text{w}}$ as well as the reset potential $V_\text{r}$, compare \cref{sec:intro:bss2}.

The \gls{bss2} operating system was used for experiment definition as well control \cite{mueller2022scalable_noeprint} and experiments are written in the \texttt{PyNN} domain specific language \cite{davison2009pynn}.

\subsection{Dataset}
In order to train the autoencoder, we create a dataset with traces recorded from the hardware.
To create a diverse set of samples, we will draw random values from a uniform distribution which covers the whole configuration range of the \gls{bss2} system for the four parameters we want to infer with the \gls{snpe} algorithm, see \cref{sec:meth:setup}.

We draw a total of \num{200000} parameterizations, emulate the model on \gls{bss2} and store the recorded membrane traces in a dataset.

\subsection{Data Embedding}\label{sec:meth:embed}
We use a convolutional autoencoder based on \cite{yildirim2018efficient} to compress our high-dimensional observation.
The model consist of several one-dimensional convolutions, ReLU activation functions, batch normalizations and max pooling layers, see \cref{tab:autoencoder}.
The input is compressed from \num{1024} data points to \num{32} data points.

\begin{table}
    \centering
	\caption{\captiontitle{Architecture of the autoencoder}
	The autoencoder is based on \cite{yildirim2018efficient} and receives an input of shape $(1,\,1024)$. 
	The batch dimension of \num{32} is omitted for readability.}
    \label{tab:autoencoder}
    \footnotesize
	\begin{tabular}{llCSS[table-format=6]}
        \toprule
		Layer & Activation & \multicolumn{1}{c}{Output  Shape} & \shortstack{Kernel \\ Shape} & \shortstack{Trainable\\ Parameters} \\
		\midrule
        \textit{Encoder} & & & & \\
		\midrule
        Conv 1D      & ReLU & (32, 1024) & 5   & 192    \\
		BatchNorm 1D & {--} & (32, 1024) &     & 64     \\
		MaxPool 1D   & {--} & (32, 512)  & 2   &        \\
        Conv 1D      & ReLU & (16, 512)  & 3   & 1552   \\
		BatchNorm 1D & {--} & (16, 512)  &     & 32     \\
		MaxPool 1D   & {--} & (16, 256)  & 2   &        \\
        Conv 1D      & ReLU & (64, 256)  & 11  & 11328  \\
		MaxPool 1D   & {--} & (64, 128)  & 2   &        \\
        Conv 1D      & ReLU & (128, 128) & 13  & 106624 \\
		MaxPool 1D   & {--} & (128, 64)  & 2   &        \\
        Conv 1D      & ReLU & (1, 64)    & 3   & 385    \\
		MaxPool 1D   & {--} & (1, 32)    & 2   &        \\
		\midrule
        \textit{Decoder} & & & & \\
		\midrule
        Conv 1D      & ReLU & (128, 32)  & 3   & 512    \\
		Upsample     & {--} & (128, 64)  &     &        \\
        Conv 1D      & ReLU & (64, 64)   & 13  & 106560 \\
		Upsample     & {--} & (64, 128)  &     &        \\
        Conv 1D      & ReLU & (16, 128)  & 11  & 11280  \\
		Upsample     & {--} & (16, 256)  &     &        \\
        Conv 1D      & ReLU & (32, 256)  & 3   & 1568   \\
		Upsample     & {--} & (32, 512)  &     &        \\
		Conv 1D      & {--} & (1, 512)   & 5   & 161    \\
		Upsample     & {--} & (1, 1024)  &     &        \\
		\midrule
        \textbf{Total} & & & & 240258  \\
		\bottomrule
    \end{tabular}
\end{table}

Before feeding the data in our network, we preprocess it.
First, we further down sample the recorded membrane traces from \num{10000} to \num{1024} data points such that they fit the input layer of our network.
Next, we normalize the recorded membrane voltages $V_\text{m}$ to be in the range from \numrange{0}{1}.
Since we use an \gls{adc} with \SI{10}{\bit} resolution to sample the membrane voltage, this can be archived by dividing all values by \num{1023}\footnote{The \gls{adc} is configured such that the maximum of the readout range is never reached.}.

The dataset is split into training, validation and test set with a ratio of $8/1/1$.
Afterwards, the training set is divided in batches of size \num{32} and the autoencoder is trained for \num{150} epochs.
The mean-squared-error between the original trace and the reconstructed trace was used as a loss function.
In each epoch, we record the loss of each batch on the training as well as the validation set.
After each epoch, we save the model parameters if the validation loss has decreased.

We used the Adam optimizer \cite{kingma2014adam} for updating the weights of the autoencoder.
During a warm-up phase, the learning rate was linearly increased from \num{1e-8} to the base learning rate of \num{1e-4} in the first \num{2000} batches.
After epoch \num{70}, the learning rate was decreased exponentially with a factor of \num{0.94}.

The model and training are implemented in the \texttt{PyTorch} library \cite{paszke2019pytorch}.

\subsection{Simulation-based Inference}
After successful training of the autoencoder, we use it in conjunction with an \gls{nde} to approximate the posterior of the parameters we altered during the generation of the dataset.
The experimental observation is fed into the encoder section of the trained autoencoder, while the NDE receives the output from its latent space.
We use a \gls{maf} with five transformations -- each transformation is made up of two blocks with \num{50} hidden units per block -- as an \gls{nde}; this \gls{nde} has been extensively used in previous publications \cite{lueckmann2021benchmarking, goncalves2020training,kaiser2023simulation}.

Similarly to the generation of the dataset, we use uniform priors over the whole parameter range as an input to the \gls{snpe} algorithm.
We train the \gls{nde} for \num{20} rounds with \num{1000} samples in each round.
The pre-trained encoder is used to reduce the dimensionality of the observed data.
During the training of the \gls{nde}, the encoder is further retrained in parallel to improve inference performence.
This transfer learning approach for the encoder is motivated by the idea that the optimal features for reconstruction may not necessarily be the most suitable for parameter inference.
At the same time, it allows for faster convergence compared to training from scratch.

We used the implementation of the \texttt{sbi} Python package \cite{tejero2020sbi} for the \gls{snpe} algorithm.
This package supports the parallel training of the autoencoder and the \gls{nde}.

\section{Results}\label{sec:results}
In the following, we will first present the generated dataset.
Next, we train an autoencoder using this dataset and evaluate how well the reconstruction agrees with the original traces.
Finally, we use the encoder part of the trained autoencoder to reduce the dimensionality of our observations and train a \gls{nde} using the \gls{snpe} algorithm.

\subsection{Dataset}

\begin{figure}
  \includegraphics[width=\linewidth]{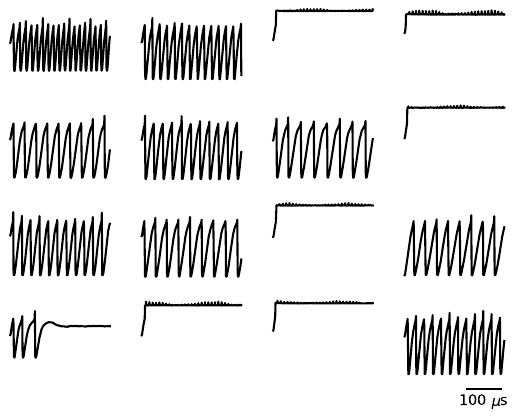}
  \caption{\captiontitle{Random samples drawn from the dataset.}
	For visualization, only the first \SI{300}{\us} are displayed; the traces are recorded for \SI{1}{\ms}.
	Due to the finite sampling frequency of the \gls{adc} and the interpolation of the recorded traces, the potentials at spike time are not identical.
	When the reset potential $V_\text{r}$ is high, the membrane voltage remains at high levels.}
  \label{fig:dataset}
\end{figure}

\Cref{fig:dataset} displays examples of traces in the dataset.
In most cases, the neuron fires over the whole displayed range.

The variation in the reset potential $V_\text{r}$ is most visible in the given example traces.
In cases where the reset voltage is high, the membrane potential remains at high values and spikes are not clearly visible.
In one case, a clear adaptation is visible: the inter-spike interval increases from spike to spike and the neuron stops firing after the third spike.

\subsection{Data Embedding}

\begin{figure}
  \includegraphics[width=\linewidth]{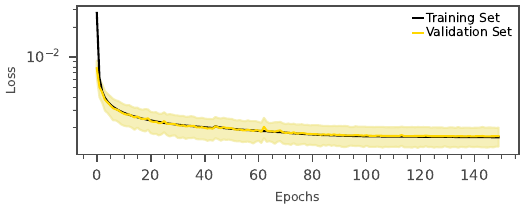}
  \caption{\captiontitle{Training of the autoencoder}
	Mean test and validation loss during training as well as one standard deviation of the validation loss.
	Both losses decrease continuously over the course of the training.}
  \label{fig:loss}
\end{figure}

We trained the autoencoder introduced in \cref{sec:meth:embed} for \num{150} epochs.
The training as well as the test loss decrease during training and start to saturate at the end of the training, \cref{fig:loss}, indicating that the chosen number of epochs is sufficient.
The difference between the two losses remains small, indicating that the model does not tend to overfit.
We recorded the lowest validation loss after \num{140} epochs and will use this model for all future evaluations.
At this point, the validation loss (\num{0.00163}) is close to the test loss (\num{0.00164}); further suggesting that our model does not overfit.

\begin{figure}
  \includegraphics[width=\linewidth]{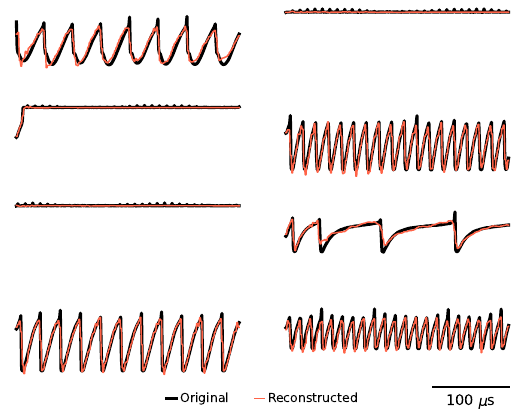}
  \caption{\captiontitle{Reconstructions of the autoencoder}
	Randomly drawn voltage traces from the test set and their reconstructions by the trained autoencoder.
	Traces are only displayed for \SI{300}{\us} to aid visual comparison.
	The reconstructions (red) follow the original traces (black) closely on a long time scale.
	However, unlike the original traces, they show some high frequency fluctuations.}
  \label{fig:reconstruction}
\end{figure}

In \cref{fig:reconstruction}, we drew random samples from the test set and fed them into the trained autoencoder.
Overall, the reconstructions follow the original samples closely.
The reconstructions of periodic membrane traces are slightly better than of traces for which the inter-spike interval changes due to adaptation.
On a fast time scale, the reconstructions show fluctuations which are not present in the original trace.

\subsection{Simulation-based Inference}
\begin{figure}
 \includegraphics[width=\linewidth]{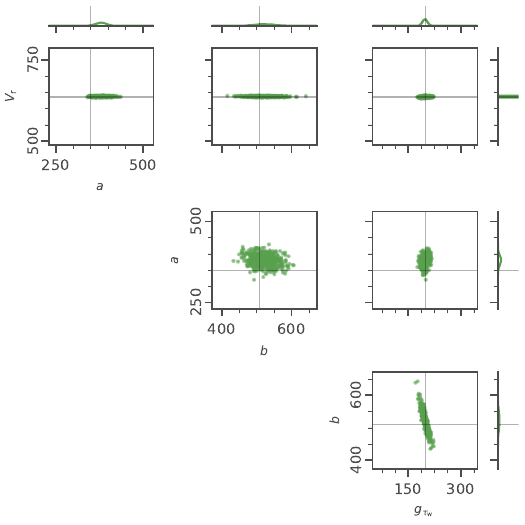}
  \caption{\captiontitle{Samples drawn from the approximated posterior}
	One- and two-dimensional marginals of \num{500} samples drawn from the approximated posterior.
	The vertical and horizontal lines represent the parameterization of the target trace.
	Note, uniform priors from \numrange{0}{1022} were chosen for all parameters, i.e.\ the posterior distribution is restricted to a much smaller region of the parameter space.}
  \label{fig:sbi}
\end{figure}

After estimating the posterior with the \gls{snpe} algorithm, we drew samples from it, \cref{fig:sbi}.
The samples are closely scattered around the parameters which were used to create the target observation.

The distribution of parameters is narrowest for the reset potential $V_\text{r}$ and the conductance $g_{\tau_\text{w}}$ controlling the adaptation time constant.
This indicates that the observations are more sensitive to these two parameters than to the subthreshold adaptation $a$ and the spike-triggered adaptation $b$.

Most of the parameters seem uncorrelated.
Only for the spike-triggered adaptation $b$ and the conductance $g_{\tau_\text{w}}$ a negative correlation can be observed: smaller values in $g_{\tau_\text{w}}$ can be compensated by higher values in $b$.
Based on the design of the circuit, a dependency of the strength of the adaptation current $w$ on the adaptation time constant $\tau_\text{w}$ is expected \cite{billaudelle2022accurate}.
This effect can be compensated by adjusting the spike-triggered adaptation $b$.

\begin{figure}
  \includegraphics[width=\linewidth]{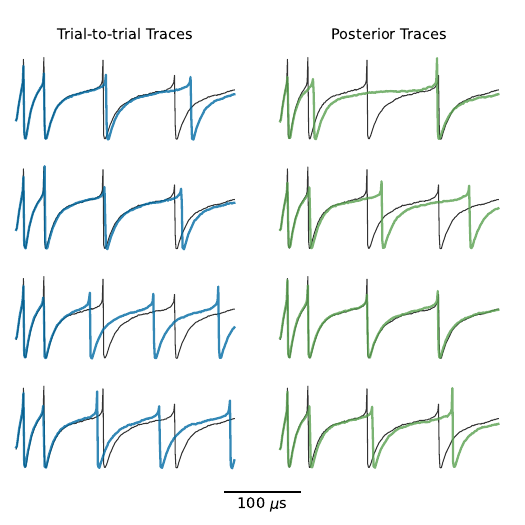}
  \caption{\captiontitle{Example traces for different experiment trials and parameterizations drawn from the approximated posterior}
	Black traces represent the chosen target observation.
	On the left side, the experiment is repeated several times with the same parameterization.
	Due to temporal fluctuations in the analog components of \acrlong{bss2}, the traces do not align exactly.
	However, the overall behavior matches between all traces.
	On the right side, we drew four parameterizations form the approximated posterior, \cref{fig:sbi}, and emulated the neuron behavior.}
  \label{fig:sbi_traces}
\end{figure}

In order to get an impression how well parameters drawn from the posterior distribution reproduce the target observation, we drew random values from the posterior, \cref{fig:sbi}, and emulated the neuron behavior with them, \cref{fig:sbi_traces}.
The recorded membrane traces match the target trace closely until the second spike.
Afterwards, the traces start to diverge from the target observation.
The divergence can to some extent be attributed to temporal noise in the analog core on \gls{bss2}.
As a comparison, \cref{fig:sbi_traces} also shows membrane recordings for the same parameters with which the target observation was recorded.
Here, the traces also start to diverge after the second spike.

In comparison to the trial-to-trial variations, the variations in the traces recorded for different parameters drawn from the posterior seem more pronounced.
Nevertheless, given the diversity seen in the traces of the dataset, \cref{fig:dataset}, the approximated posterior seems to yield parameters which closely resemble the given target observation.

\section{Discussion}\label{sec:discussion}
\glsresetall

In this study, we demonstrated that the \gls{snpe} algorithm can be successfully leveraged to infer model parameters for the \gls{adex} neuron model emulated on the neuromorphic \gls{bss2} system when utilizing membrane recordings as observations.
Despite the presence of temporal noise on the hardware, the algorithm was able to approximate the posterior distribution, thereby enabling adequate emulation of the target trace.

We began by generating a dataset consisting of diverse voltage traces.
After initial preprocessing of the data, a convolutional autoencoder was successfully trained, whose encoder was then used to compress the time series data into just 32 features.
Subsequently, this lower-dimensional representation of the data was fed into the \gls{nde} of the \gls{snpe} algorithm.
To further enhance inference performance, we chose to simultaneously retrain the pretrained encoder alongside the \gls{nde}.
Our method yielded promising results, as the algorithm was able to identify the correct region within the 4-dimensional parameter space, even in the presence of the trial-to-trial variations inherent to the hardware.

The reconstructions of the autoencoder could have been improved by selecting a larger latent space dimension.
However, a balance must be struck between reconstruction accuracy and dimensionality reduction for the \gls{snpe} algorithm.
As shown in \cref{fig:reconstruction}, the reconstructions of traces with stronger adaptation are worse than those of other traces.
This is due to the low occurrence of adaptation traces in the training set --- only a small subset of the chosen parameter space produces such traces.
Since our observation was an adaptation trace, this reinforces our decision to simultaneously retrain the encoder with the \gls{nde}.

Temporal noise in the analog components leads to the approximation of the posterior being broader, as parameter values that would typically produce a different voltage trace might resemble one that closely matches the initial target trace.
Thus, these values are interpreted as if they produce results similar to those of the true parameters.
At the same time, the target trace itself could represent a variation of the typical trace at the target parameters.
This would explain the slight offset between the peak of the posterior distribution and the true parameter values, as displayed in \cref{fig:sbi}.

For $g_{\tau_\text{w}}$ and $V_\text{r}$, more narrow marginals could be identified than for the other two parameters, see \cref{fig:sbi}. 
However, a broader posterior does not necessarily indicate a worse performance of the approximation method, as certain variations in a parameter may not significantly impact the resulting trace.
Furthermore, compensation mechanisms between parameters might exist.
For instance, such a relationship was particularly evident between the spike-triggered adaptation $b$ and the conductance $g_{\tau_\text{w}}$, exhibiting a negative correlation due to the specific circuit design \cite{billaudelle2022accurate}.

Finally, taking trial-to-trial variations into account, most of the emulated traces of randomly sampled posterior values closely resemble the original target trace, as shown in \cref{fig:sbi_traces}.
Hence, our method utilizing the autoencoder proved effective and eliminated the reliance on handcrafted features for the dimensionality reduction of the traces.

The future objective of this work is to develop an inference pipeline for membrane voltage recordings from biological neurons to enable precise emulation.
However, this requires the accurate inference of more parameters of the \gls{bss2} system \cite{pehle2022brainscales2}.
To achieve this, several optimizations of our method can be explored.

First, extensive testing and hyperparameter searches for the \gls{nde} in higher-dimensional parameter spaces are needed to assess the robustness of our current method.

Furthermore, different autoencoder architectures could be explored to improve compression efficiency.
The input size of the network could be increased too, allowing it to handle traces emulated for longer durations.

In addition, approaches to address temporal noise could be considered to potentially achieve a more accurate posterior, as this noise also complicates posterior analysis.
Building on this, ensembles comprising multiple posteriors from various emulations of the true parameters might counteract overconfident posterior estimates \cite{hermans2022crisis}.

Finally, the posterior distribution needs to undergo more rigorous testing.
A suitable metric for measuring the similarity between different voltage traces must be established, such that systematic posterior-predictive checks can be performed \cite{hermans2022crisis}.
These would filter out posterior estimates that do not align well with the observation.
Moreover, the posterior could be used to explore further correlations between different parameters.
Additionally, sensitivity analysis could help identify critical directions in the parameter space where changes in the parameters have a strong impact on the resulting trace~\cite{constantine2014active}.

To date, studies have primarily applied the \gls{snpe} algorithm to reproduce the behavior of spiking neurons in numerical simulations \cite{goncalves2020training, deistler2022truncated, lueckmann2017flexible} or of passive neurons on neuromorphic hardware \cite{kaiser2023simulation}.
Thus, this work represents a novel approach and an initial step towards applying the algorithm to reproduce biological neuron behavior on neuromorphic hardware.
Fast emulation on accelerated neuromorphic hardware has the potential to explore new questions that are difficult to address by slower numerical simulation techniques~\cite{zenke2014limits}.
Consequently, the combination of automatic feature extraction with the application of the \gls{snpe} algorithm for emulation on neuromorphic hardware could contribute to shaping future methodologies in neuroscience research.

\section*{Acknowledgements}
\addcontentsline{toc}{section}{Acknowledgment}

This research has received funding from
the European Union's
Horizon 2020 research and innovation programme under grant agreement No.\ 
945539 (Human Brain Project SGA3)
and Horizon Europe grant agreement No.\ 101147319 (EBRAINS 2.0),
and the \foreignlanguage{ngerman}{Deutsche Forschungsgemeinschaft} (DFG, German Research Foundation) under Germany's Excellence Strategy EX 2181/1-390900948 (the Heidelberg \mbox{STRUCTURES} Excellence Cluster).

\section*{Author Contributions}\label{sec:author_contributions}

We give contributions in the \textit{CRediT} (Contributor Roles Taxonomy) format:
\textbf{JH}: Investigation, visualization, methodology, software;
\textbf{JK}: Conceptualization, methodology, supervision, software, visualization;
\textbf{EM}: Conceptualization, methodology, supervision, software, resources;
\textbf{JS}: Conceptualization, methodology, supervision, funding acquisition;
\textbf{all}: writing --- original draft, writing --- reviewing \& editing.

\printbibliography

\end{document}